\documentclass[english]{article}
\usepackage{ae,aecompl}
\usepackage[T1]{fontenc}
\usepackage[latin9]{inputenc}
\usepackage{geometry}
\usepackage{array}
\usepackage{float}
\usepackage{booktabs}
\usepackage{url}
\usepackage{amsmath}
\usepackage{graphicx}
\usepackage{setspace}
\makeatletter

\providecommand{\tabularnewline}{\\}
\floatstyle{ruled}
\newfloat{algorithm}{tbp}{loa}
\providecommand{\algorithmname}{Algorithm}
\floatname{algorithm}{\protect\algorithmname}

\usepackage{algorithm, algpseudocode}
\date{}
\usepackage{manyfoot}
\DeclareNewFootnote{A}[roman]

\@ifundefined{showcaptionsetup}{}{%
 \PassOptionsToPackage{caption=false}{subfig}}
\usepackage{subfig}
\makeatother

\usepackage{babel}
\begin{document}

\title{Multiple sclerosis lesion enhancement and white matter region estimation
using hyperintensities in FLAIR images}

\author{Paulo G. de Lima Freire$^{a}$\thanks{\protect\url{paulo.freire@ufscar.br}, corresponding author},
Ricardo J. Ferrari$^{a}$\thanks{\protect\url{rferrari@ufscar.br}}}
\maketitle
\begin{center}
{\footnotesize{}$^{a}$}{\small{}Departamento de Computação, Universidade
Federal de São Carlos, Rod. Washington Luís, Km 235, 13565-905, São
Carlos, SP, Brazil}{\small\par}
\par\end{center}
\begin{abstract}
Multiple sclerosis (MS) is a demyelinating disease that affects more
than 2 million people worldwide. The most used imaging technique to
help in its diagnosis and follow-up is magnetic resonance imaging
(MRI). Fluid Attenuated Inversion Recovery (FLAIR) images are usually
acquired in the context of MS because lesions often appear hyperintense
in this particular image weight, making it easier for physicians to
identify them. Though lesions have a bright intensity profile, it
may overlap with white matter (WM) and gray matter (GM) tissues, posing
difficulties to be accurately segmented. In this sense, we propose
a lesion enhancement technique to dim down WM and GM regions and highlight
hyperintensities, making them much more distinguishable than other
tissues. We applied our technique to the ISBI 2015 MS Lesion Segmentation
Challenge and took the average gray level intensity of MS lesions,
WM and GM on FLAIR and enhanced images. The lesion intensity profile
in FLAIR was on average 25\% and 19\% brighter than white matter and
gray matter, respectively; comparatively, the same profile in our
enhanced images was on average 444.57\% and 264.88\% brighter. Such
results mean a significant improvement on the intensity distinction
among these three clusters, which may come as aid both for experts
and automated techniques. Moreover, a byproduct of our proposal is
that the enhancement can be used to automatically estimate a mask
encompassing WM and MS lesions, which may be useful for brain tissue
volume assessment and improve MS lesion segmentation accuracy in future
works.
\end{abstract}
\textbf{Keywords:} multiple sclerosis, lesion enhancement, white matter
tissue, FLAIR images, hyperintensities

\section{\label{sec:Introduction}Introduction}

Multiple sclerosis (MS) is a demyelinating disease that attacks the
central nervous system (CNS) and affects more than 2 million people
worldwide \cite{Browne-2013}. It destroys neurons' myelin sheaths,
causing many effects on one's body such as dizziness, confusion, memory
problems and numbness of arms and legs \cite{WHOAtlasMS-2008}. The
cause of MS is still unknown, and the disease itself has a devastating
effect both for individuals and society. Since the onset of MS is
typically around age 30, it affects subjects at the peak of their
productivity in life \cite{MS-2001}. In this context, it is essential
to offer tools to neurologists and radiologists to quickly identify
the disease and prescribe treatments to help patients lead a normal
life.

Magnetic resonance imaging (MRI) is often used in diagnosis and follow-up
of MS due to its high contrast between soft tissues \cite{Compston-2008}.
One widely used imaging protocol of MRI is the Fluid Attenuated Inversion
Recovery (FLAIR), which, as the name suggests, attenuates the effect
of fluids, mainly from the cerebral spinal fluid (CSF) region. FLAIR
images are important in the MS context because MS lesions appear hyperintense
in this particular image weight, thus making it easier for physicians
to identify them \cite{Hashemi-1995}.

Though MS lesions present a hyperintense profile in FLAIR images,
their intensity range varies significantly among different patients
and between various time points from the same subject \cite{Freire-2016}.
Due to this, both manual and automatic segmentation may be affected
and undermine the accuracy of MS lesion detection, since they can
be mistaken by other brain tissues - namely, white matter (WM) and
gray matter (GM). And while it is true that gray level intensity is
not the only feature that helps experts and algorithms differ MS lesions
from other tissue classes, a more sharp distinction between hyperintensities
and normal tissues would provide extra leverage to separate each class
more accurately.

A proper distinction between brain tissues and abnormalities, such
as lesions, may be of great help for experts and automatic segmentation
techniques alike. In \cite{Shah-2011}, the authors made a comprehensive
analysis of the importance of intensity normalization and its effect
on MS lesion segmentation. They have shown that applying the intensity
normalization technique proposed by \cite{Nyul-2000} to 21 images
from subjects with MS increased the Dice Similarity Coefficient (DSC)
\cite{Dice-1945} of lesion segmentation on automatic supervised approaches.
However, a scatter plot analysis showed that despite normalization,
lesions still had a significant intensity overlap with WM and GM tissues.

In \cite{Tomas-Fernandez-2015}, the authors proposed an algorithm
to increase automatic lesion and brain tissue segmentation robustness
by estimating a spatially global within-the-subject intensity distribution
and a spatially local intensity distribution derived from a healthy
reference population. This approach tried to circumvent the overlap
between the whole brain signal intensity distribution of lesions and
healthy tissue. A scatter plot analysis showed that local intensities
from a reference population offered a better lesion intensity separation
than by either global or local intensity distributions derived from
a patient with MS. Though the authors' proposal performed better than
six other segmentation techniques on a 31-subject database, it was
still dependent on a healthy reference population and, consequently,
on careful registration and intensity normalization steps across the
healthy and lesion image sets.

In this sense, we propose a technique to enhance hyperintensities
in FLAIR images to better distinguish MS lesions from WM and GM. We
built on the works of \cite{Roy-2013,Roy-2014} to automatically generate
an image that shows the probability of each voxel being a hyperintense
one, herein called the hyperintensity map. The main advantage of this
technique is that it requires only FLAIR images and enhances MS lesions
such that their intensity profile is much brighter than WM and GM
compared to their profiles in FLAIR itself.

Regarding the effects that brain abnormalities, such as lesions, have
on brain tissue segmentation, some works in the literature have explored
the issue. In \cite{Battaglini-2012}, the authors have shown that
segmentation-based methods for brain volume measurement suffer in
the presence of lesions since they interfere with GM and WM depending
on lesion size and intensity. To overcome this problem, they proposed
filling lesions with intensities matching surrounding normal-appearing
WM. Their approach helped reduce the impact lesions had on tissue
segmentation, especially regarding GM, and improved the accuracy of
tissue classification and brain volume measurement. However, the filling
algorithm depends on having the lesion ground truths at hand and,
as noted by the authors themselves, their approach may overestimate
lesion holes depending on where they are located and the intensity
variation in the surrounding area.

Similarly, in \cite{Valverde-2015}, the authors stated that the accuracy
of automatic tissue segmentation methods may be affected by the presence
of MS lesions during the tissue segmentation process. They applied
six well-known segmentation techniques to 30 T1-weighted images from
subjects with MS and verified that GM volume was overestimated by
all methods when lesion volume increased. This overestimation persisted
even when masking out or relabeling lesions during segmentation. This
particular finding was significant because provided evidence to show
that tissue atrophy measurements are likely to be distorted when the
subject's lesion load is high.

In \cite{Valverde-2015b}, the authors conducted a study to verify
the effect ground truth annotations had on the assessment of automatic
brain tissue segmentation accuracy. More specifically, they applied
ten different brain tissue segmentation methods to the Internet Brain
Segmentation Repository (IBSR)\footnote{\url{https://www.nitrc.org/projects/ibsr}},
since this dataset considered Sulcal Cerebrospinal Fluid (SCSF) voxels
as gray matter. Though this dataset comprised only images from healthy
subjects, the authors were able to check that the performance and
accuracy of the methods on IBSR images varied significantly when not
considering SCSF voxels. This finding indicates that not only abnormalities,
such as lesions, may affect segmentation techniques, but labeling
is also a concern and may lead to a misguided analysis of accuracy.

Finally, in \cite{Valverde-2016}, the authors proposed an automated
T1-weighted/FLAIR tissue segmentation approach designed to deal with
images from subjects with WM lesions. They suggested a partial volume
tissue segmentation with WM outlier rejection and filling, along with
intensity, probabilistic and morphological prior maps, to segment
brain tissues. This approach did not need manual annotations of lesions,
which is an advantage compared to other works that use lesion filling
or masking based on expert ground truths. The authors applied their
algorithm to two databases. One of them comprised only images from
subjects with MS, and the author's proposal achieved competitive results
compared to other five segmentation techniques. However, the MS database
was not publicly available, thus making it difficult to directly compare
their results to other works in the literature.

To this end, a byproduct of our technique is the estimation of a white
matter mask based on the hyperintensity map. This approach is relevant
because, as previously noted, MS lesions may interfere with the segmentation
of brain tissues due to similarities in their intensity profiles.
Our white matter mask estimation relies on the hyperintensity map
to fill lesion holes that were left out during an automatic segmentation
process. Our technique does not require manual annotations, making
sole use of FLAIR images and probabilistic anatomical atlases to get
such estimation. To verify its accuracy, we extracted DSC and the
percentage of MS lesions that were included in the WM mask during
the process to confirm how well our approach was to get a reasonable
WM region estimate.

This paper is divided as follows. In Section \ref{sec:Materials-and-methods},
we describe our methodology and the database we used to apply our
technique on; we present our results in Section \ref{sec:Results}
and discuss them in Section \ref{sec:Discussion}. Finally, we present
our final considerations in Section \ref{sec:Conclusions} and indicate
our future works.

\section{\label{sec:Materials-and-methods}Materials and methods}

In this section, we describe the databases we used, the pipeline for
enhancing multiple sclerosis lesions in FLAIR images and the algorithm
for estimating the white matter region mask.

\subsection{\label{subsec:Database}Databases}

\subsubsection{\label{subsec:Clinical-images}Clinical images}

We used the training dataset of the Longitudinal MS Lesion Segmentation
Challenge\footnote{\url{http://iacl.ece.jhu.edu/index.php/MSChallenge/data}}
made available during the 2015 International Symposium on Biomedical
Imaging \cite{Carass-2017}. This dataset comprised images of five
patients, one male and four females, with a total of 21 time-points.
The mean age of the patients was 43.5 years and the mean time between
follow-up scans was one year. 

Each scan was imaged and pre-processed in the same manner, with data
acquired on a 3 Tesla MRI scanner (Philips Medical Systems, Best,
The Netherlands). The imaging sequences were adjusted to produce T1-weighted,
T2-weighted, proton density (PD) and FLAIR images.

Each subject underwent the following pre-processing: the baseline
(first time-point) magnetization prepared rapid gradient echo (MPRAGE)
was inhomogeneity-corrected using N4 \cite{Tustison-2009}, skull-stripped
\cite{Carass-2007,Carass-2011} and dura stripped \cite{Shiee-2014},
followed by a second N4 inhomogeneity correction and rigid registration
to a 1 mm isotropic MNI template. Once the baseline MPRAGE was in
MNI space, it was used as a target for the remaining images, which
included the baseline T2-w, PD-w, and FLAIR, as well as the scans
from each of the follow-up time-points. These images were then N4
corrected and rigidly registered to the 1 mm isotropic baseline MPRAGE
in MNI space. In the end, image dimensions were $181\times217\times181$.

It is important to note that the training dataset also included manual
MS lesion delineations by two experts for each time-point. More details
about time-points and average MS lesion volume for each subject are
summarized in Table \ref{tab:Patients}.

\begin{table}[tbh]
\begin{centering}
\begin{tabular}{c>{\centering}p{2cm}>{\centering}p{3cm}>{\centering}p{3cm}}
\toprule 
 & Number of time-points & Mean lesion volume (in ml)

Expert 1 & Mean lesion volume (in ml)

Expert 2\tabularnewline
\midrule
\midrule 
Patient 1 & 4 & 16.67 & 19.07\tabularnewline
\midrule
\midrule 
Patient 2 & 4 & 30.52 & 31.80\tabularnewline
\midrule
\midrule 
Patient 3 & 5 & 5.40 & 7.81\tabularnewline
\midrule
\midrule 
Patient 4 & 4 & 2.17 & 3.23\tabularnewline
\midrule
\midrule 
Patient 5 & 4 & 4.55 & 3.96\tabularnewline
\bottomrule
\end{tabular}
\par\end{centering}
\caption{\label{tab:Patients}Number of time-points and average lesion volume
for each patient.}
\end{table}

\subsubsection{\label{subsec:Probabilistic-atlases}Probabilistic anatomical atlases}

To estimate white matter regions and identify gray matter clusters,
we used the probabilistic atlases from the ICBM project \cite{Fonov-2009}.
The atlases spatial resolution was $1\times1\times1$ mm and their
initial dimensions were $256\times256\times256$. However, they were
registered to each clinical time-point to provide accurate spatial
information. A T1-weighted image initially registered to both white
matter and gray matter atlases was used as a moving image, while T1-weighted
images from each time-point were used as reference images. The registration
took place using the NiftyReg tool \cite{Modat-2010} with free-form
B-Spline deformation model and multi-resolution approach for non-rigid
registration. The transformation was then applied to the atlases,
thus making them aligned to each time-point and with dimensions $181\times217\times181$. 

\subsection{\label{subsec:Metrics}Metrics}

A total of three metrics were used to assess the intensity profile
distinction between lesions and other brain tissues and to compare
the estimated white matter mask with the ground truth.

The intensity profile distinction (IPD) is calculated as 
\begin{equation}
IPD=\left(\frac{\text{average(lesions)}}{\text{average(tissue)}}-1\right)\times100,\label{eq:IPD}
\end{equation}
in which we simply divide the average intensity of lesions by the
average intensity of a tissue of interest (white matter or gray matter)
in a particular image and then scale it in terms of percentage. By
doing so, we can verify how much brighter, percent-wise, the lesion
cluster is than other tissues.

On the assessment of the estimated white matter mask, we used the
Dice Similarity Coefficient (DSC) \cite{Dice-1945} to check the overall
overlap between our estimation and the ground truth. The DSC is defined
as 
\begin{equation}
DSC=\frac{2\times TP}{FP+FN+2\times TP},\label{eq:DSC}
\end{equation}
where TP, FP, and FN are the true positives, false positives and false
negatives. DSC values fall in the interval $[0,1]$, and the closer
they are to 1, the better.

Another metric we used to assess the white matter mask estimation
was the lesion intersection (LI), defined as 
\begin{equation}
LI=\frac{|\text{Lesion}_{GT}\cap\text{Mask}_{estim}|}{|\text{Lesion}_{GT}|}\times100.\label{eq:LI}
\end{equation}

Similar to IPD, LI is also calculated in terms of percentage and provides
a quantitative tool to analyze the lesion load that was kept during
the white matter mask estimation.

\subsection{\label{subsec:Pre-processing}Pre-processing}

The pre-processing stage used in this work was comprised of three
steps: noise reduction, intensity normalization and intermediate enhancement.

The noise reduction step was performed using the non-local means approach
\cite{Buades-2005} with $\sigma=15$. Reducing image noise is important
because it helps eliminate part of intensity variations within the
image, creating a smoother profile for each brain tissue. Given the
nature of our lesion enhancement technique, which is heavily based
on gray level intensities, noise reduction is relevant to mitigate
effects inherent in the image acquisition procedure and thus improve
the enhancement of MS lesions while dimming out other brain tissues.

Intensity normalization was done in order to assure that every image
had the same intensity range. We applied the normalization proposed
in \cite{Roy-2013}, which rescales intensities as $v'=\frac{v}{\mu+3\sigma}$,
where $v'$ is the new value of a given voxel, $v$ is the voxel value
in FLAIR and $\mu$ and $\sigma$ represent the mean and standard
deviation of the whole brain, respectively.

An edge detection step using Sobel \cite{Duda-2000} is required as
input to generate the intermediate enhanced image with increased contrast
between MS lesions and their surroundings, namely white matter tissue.
Following the proposal in \cite{Roy-2013}, this intermediate image
was created as follows. Let $s=\{x,y,z\}$ be a particular spatial
location, $I_{s}$ the FLAIR intensity at $s$ and $g_{s}$ the gradient
in the Sobel image at $s$. The edge and intensity information are
combined as 
\begin{equation}
h(i)=\frac{1}{N}\sum_{s\in\{s|I_{s}=i\}}Prob\left(g\leq g_{s}\right),\label{eq:histogram}
\end{equation}
 where $N$ is the total number of voxels with intensity $i$. In
other words, Equation \ref{eq:histogram} goes over every FLAIR image
intensity $i$ and, given the probability density function (PDF) of
the Sobel image, sums up the histogram bins that have a smaller frequency
than $g_{s}$ and then normalizes it.

\begin{figure}[h]
\noindent \begin{centering}
\subfloat[]{\begin{centering}
\includegraphics[scale=0.6]{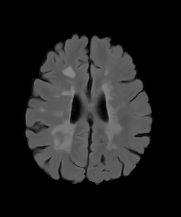}
\par\end{centering}

}\subfloat[]{\begin{centering}
\includegraphics[scale=0.6]{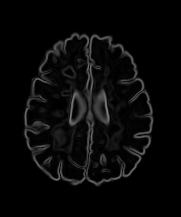}
\par\end{centering}
}\subfloat[]{\begin{centering}
\includegraphics[scale=0.6]{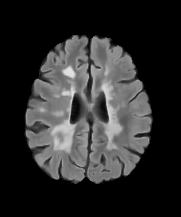}
\par\end{centering}
}
\par\end{centering}
\caption{\label{fig:Intermediate-image}Intermediate image generation. (a)
FLAIR image, (b) Sobel image, (c) intermediate image.}

\end{figure}

After calculating $h(i)$, we may compute the cumulative distribution
function (CDF) of $h$ as proposed in \cite{Roy-2013}:
\begin{equation}
q(i)=\sum_{k=1}^{i}h(k).
\end{equation}
In the end, each $q(i)$ is used to replace each intensity $i$. An
example of such intermediate image is shown in Figure \ref{fig:Intermediate-image}.

\subsection{\label{subsec:Hyperintensity-map}Hyperintensity probability map}

The hyperintensity probability map is calculated based on the intermediate
image generated during the pre-processing stage detailed in Section
\ref{subsec:Pre-processing}. We devised an algorithm that automatically
generates such map and does not depend on parameters that must be
set by experimental observations, as opposed to \cite{Roy-2013,Roy-2014}. 

The central principle behind this map is to compare each voxel neighborhood
intensity with patches across different points in the image. The more
times the voxel's neighborhood mean intensity is higher than the patches',
then the more likely it is for that particular voxel to stand out
and more likely for it to have a high hyperintense probability.

The first step to create the map is to define where each patch will
be centered. To do that, we define a point net for each slice following
the algorithm proposed in \cite{Roy-2014}, which uses the combination
of sines and cosines to evenly distribute points across a slice. Let
$p=\{x,y,z\}$ be the coordinates of a candidate point. We then create
new points $p'=\{x',y',z\}$ with
\begin{equation}
\begin{array}{cc}
x'= & x+r\cos\theta\\
y'= & y+r\sin\theta
\end{array},\label{eq:sine-cosine}
\end{equation}
where $\theta$ is the angle and $r=10$ is the radius. We set $\theta$
to zero and increase it by 60 degrees six times to complete a whole
circumference. The six newly defined points become candidate points,
and the process repeats itself until no new point is found.

\begin{figure}[h]
\begin{centering}
\subfloat[]{\begin{centering}
\includegraphics[scale=0.6]{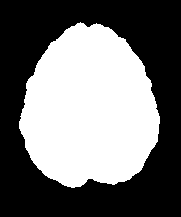}
\par\end{centering}

}\subfloat[]{\begin{centering}
\includegraphics[scale=0.6]{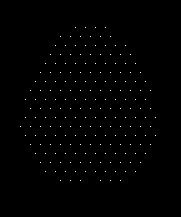}
\par\end{centering}

}
\par\end{centering}
\caption{\label{fig:Point-net}Brain mask (a) and pruned point net (b). Each
patch of this particular slice is centered around one point in (b).}
\end{figure}

After defining such points, we prune the net using a brain mask to
only keep points that are inside our ROI. This procedure is done by
simply purging points outside the brain mask. An example of a final
point net set $P$ for a particular slice is shown in Figure \ref{fig:Point-net}.

Now, let $\mu$ and $\sigma$ be the mean and standard deviation of
the whole intermediate image within the brain mask. Then, for each
voxel, we calculate its neighborhood mean intensity as

\begin{equation}
\mu_{v}=\frac{1}{N_{v}}\sum_{k=1}^{N_{v}}i_{k},\label{eq:mean_neighbor}
\end{equation}
where $\mu_{v}$ is the mean neighborhood intensity of voxel $v$,
$N_{v}$ is the number of neighbors of $v$ and $i_{k}$ is the intensity
of neighbor $k$. The neighborhood size was defined as $3\times3\times3$
in order to maintain a good trade-off between sharpness and smoothness.
The same rationale is used for the patches: the mean intensity is
calculated as in Equation \ref{eq:mean_neighbor}, thus creating $\mu_{p}$
for each patch.

Finally, we create a score $S_{v}$ as
\begin{equation}
S_{v}=\frac{1}{|P|}\sum_{p\in P}\delta(\mu_{v},\mu_{p}),\label{eq:score}
\end{equation}
where $|P|$ is the cardinality of the patch set and
\begin{equation}
\delta(\mu_{v},\mu_{p})=\begin{cases}
1, & \text{if }\mu_{v}-\mu_{p}\geq\sigma\\
0, & \text{otherwise}
\end{cases}.\label{eq:delta}
\end{equation}

\begin{figure}[h]
\begin{centering}
\subfloat[]{\begin{centering}
\includegraphics[scale=0.6]{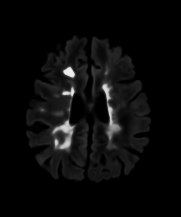}
\par\end{centering}
}\subfloat[]{\begin{centering}
\includegraphics[scale=0.6]{t02c01_flair}
\par\end{centering}
}\subfloat[]{\begin{centering}
\includegraphics[scale=0.6]{t02c01_enhanced}
\par\end{centering}
}
\par\end{centering}
\caption{\label{fig:Hyperintensity-map}Hyperintensity map (a) compared to
the original FLAIR image (b) and the intermediate image (c).}
\end{figure}

In other words, if the difference of intensity between a candidate
voxel neighborhood and a patch is greater than or equal to the standard
deviation of the whole image, then it is a hit. Otherwise, it is a
miss. By doing so, voxels with bright neighborhoods are enhanced while
other regions and tissues are dimmed out. Moreover, since we normalize
the score $S_{v}$, each voxel remains in the range $[0,1]$, which
also serves as a hyperintensity probability indicator. An example
of the map is shown in Figure \ref{fig:Hyperintensity-map}.

It is important to note that our approach does not require a hard
threshold for Equation \ref{eq:delta} or a fixed number of patches
as it happens in \cite{Roy-2013}. Instead, the threshold is calculated
automatically with respect to the standard deviation. Though simple,
this is a significant improvement, since the main problem of using
a hard threshold is that even normalized, intensities inherently vary
from image to image. In this sense, a soft threshold such as the one
we propose in this work may offer a better option for the enhancement
of hyperintensities for it can adapt to each image intensity profile.

\subsection{\label{subsec:White-matter-mask}White matter mask estimation}

The white matter region usually comprises most MS lesions \cite{Compston-2008}.
An automatic brain segmentation into three clusters (WM, GM, and CSF)
based on gray level intensities is most certainly going to mix lesions
and cluster them as GM, WM or both \cite{Battaglini-2012,Valverde-2015}.
In this context, being able to estimate a mask that encompasses both
white matter tissue and MS lesions may narrow down the ROI and increase
the accuracy of lesion segmentation. To do so, we leveraged the fact
that the map described in Section \ref{subsec:Hyperintensity-map}
can also be interpreted as a probability map and used it to get an
estimate of such mask. 

In this work, we made use of the Student t mixture model proposed
in \cite{Freire-2016} and used T1-weighted and FLAIR images from
each time-point to segment the brain into three different clusters
and get an initial WM mask, herein referred to as $WM_{initial}$.
To automatically identify the WM cluster from others, we used the
WM probability map described in Section \ref{subsec:Probabilistic-atlases},
averaged it over each cluster and selected the one with the highest
WM probability.

Since the 2015 Longitudinal MS Lesion Segmentation Challenge did not
provide WM ground truths, we created our own using a straightforward
approach. Given any $WM_{initial}$, we simply merged it with the
lesions ground truth to get the whole WM region in one single mask,
herein referred to as $WM_{whole}$. Considering that each time-point
had two different lesions ground truth, we created two WM masks for
each time-point as well. 

\begin{figure}[!tbph]
\begin{centering}
\subfloat[]{\begin{centering}
\includegraphics[scale=0.45]{t02c01_contrast}
\par\end{centering}

}\subfloat[]{\begin{centering}
\includegraphics[scale=0.45]{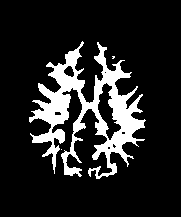}
\par\end{centering}
}\subfloat[]{\begin{centering}
\includegraphics[scale=0.45]{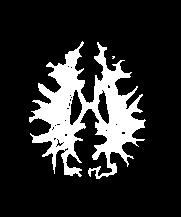}
\par\end{centering}
}\subfloat[]{\begin{centering}
\includegraphics[scale=0.45]{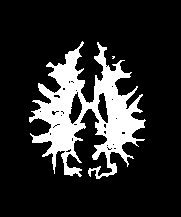}
\par\end{centering}
}\caption{\label{fig:White-matter-estimation-output}White matter estimation
output. (a) Hyperintensity map, (b) $WM_{initial}$, (c) $WM_{ground\,truth}$
and (d) $WM_{estim}$.}
\par\end{centering}
\end{figure}

The actual WM estimation took place as follows. First, we calculated
the mean ($\mu_{HI}$) and standard deviation ($\sigma_{HI}$) of
the region defined by $WM_{initial}$ on the hyperintensity map image
and the mean ($\mu_{prob}$) of the region defined by $WM_{initial}$
on the WM probability atlas. The idea was to expand $WM_{initial}$
by considering voxels that are not part of the mask yet and analyzing
$3\times3\times3$ neighborhoods centered around these voxels to verify
their potential for being included. The expansion itself occurred
by incorporating voxels that seemed as outliers; more precisely, voxels
with mean neighborhood values greater than $\mu_{HI}+1\times\sigma_{HI}$
in the hyperintensity map and greater than $\mu_{prob}$ in the probability
atlas. A pseudo-algorithm for estimating the white matter mask is
presented in Algorithm \ref{alg:WM-estimation} and an example of
the output of this estimation is shown in Figure \ref{fig:White-matter-estimation-output}.

\begin{algorithm}[h]
Input: $\text{WM}_{initial}$, $\text{HI}_{map}$, $\text{WM}_{prob}$

Output: $\text{WM}_{estim}$

\begin{algorithmic}[1]
\State{$i\gets 3$}
\State{$WM_{estim}\gets WM_{initial}$}
\State{$\mu_{HI}, \sigma_{HI} \gets MeanAndSigma(WM_{estim}, HI_{map})$}
\State{$\mu_{prob} \gets Mean(WM_{estim}, WM_{prob})$}
\State{$t_{prob}\gets \mu_{prob}$}
\State{$t_{HI} \gets \mu_{HI} + i\times\sigma_{HI}$}
\For{each voxel \textbf{not} in $WM_{estim}$}
	\State{$WM_{estim} \gets ExpandWM(WM_{estim}, HI_{map}, WM_{prob}, t_{HI}, t_{prob}) $}
\EndFor
\end{algorithmic}

\caption{\label{alg:WM-estimation}White matter mask estimation algorithm}

\end{algorithm}

\subsection{Pure white matter and gray matter clusters}

To estimate the intensity profiles of white matter and gray matter
clusters without lesions, we did the following. For each time-point
segmentation, we automatically identified the white and gray matter
clusters by analyzing their mean intensities on white and gray matter
probabilistic atlases. The cluster with highest white matter atlas
mean intensity was taken as the white matter cluster ($WM_{initial}$);
the same rationale was used for the gray matter cluster ($GM_{initial}$).
Then, for each expert annotation, we simply excluded every voxel that
had any intersection with the lesion ground truth and these two clusters.
Formally, $WM_{pure}=WM_{initial}-(WM_{initial}\cap GT_{E\in\{\text{Expert 1, Expert 2}\}})$
and $GM_{pure}=GM_{initial}-(GM_{initial}\cap GT_{E\in\{\text{Expert 1, Expert 2}\}})$.
By doing so, we were able to get so-called ``pure'' WM and GM clusters,
which were then used to calculate their intensity profiles and compare
them to lesion profiles in Section \ref{subsec:Brightness-profile}.

\section{\label{sec:Results}Results}

In this section, we present the results regarding the brightness intensity
profile of MS lesions compared to gray matter and white matter on
FLAIR, intermediate and hyperintensity map images for each patient.

For the sake of comparison, all images were rescaled to the range
$[0,1]$ and the results in Section \ref{subsec:Brightness-profile}
are shown in percentage; that is, how much brighter, percent-wise,
the lesion profile was compared to other brain tissues.

We also present the Dice and lesion intersection metrics regarding
the white matter mask in Section \ref{subsec:White-matter-comparison}
to offer a quantitative analysis of the mask estimation. 

It is important to remember that each patient had a number of time-points
made available. For the sake of understandability and inter-patient
comparison, every result presented in Sections \ref{subsec:Brightness-profile}
and \ref{subsec:White-matter-comparison} represents the average of
all time-points of a particular patient.

\subsection{\label{subsec:Brightness-profile}Brightness profile}

\begin{figure}[!tbph]
\begin{centering}
\subfloat[]{\begin{centering}
\includegraphics[scale=0.46]{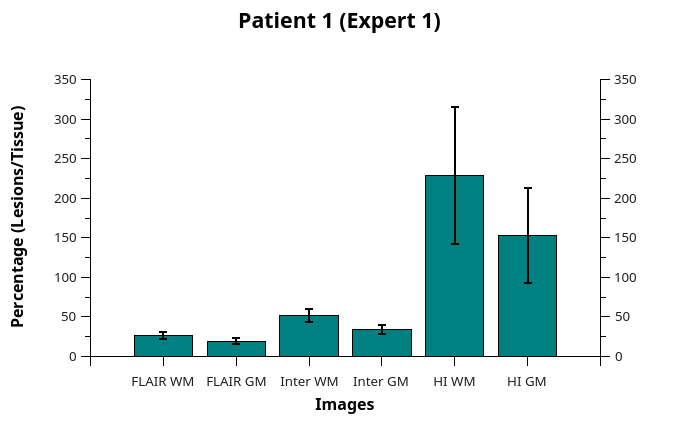}
\par\end{centering}
}\subfloat[]{\centering{}\includegraphics[scale=0.46]{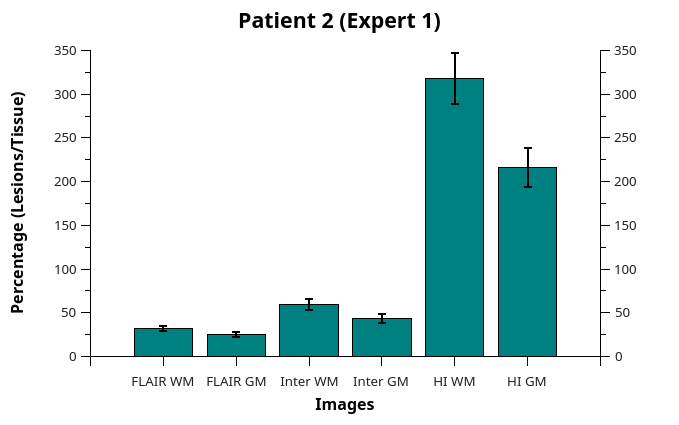}}
\par\end{centering}
\begin{centering}
\subfloat[]{\centering{}\includegraphics[scale=0.46]{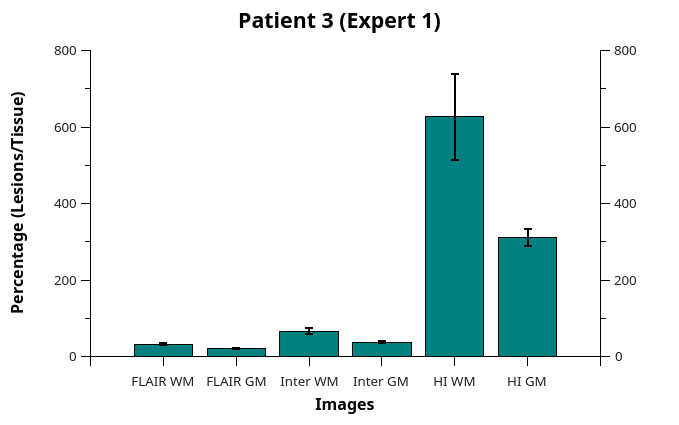}}\subfloat[]{\centering{}\includegraphics[scale=0.46]{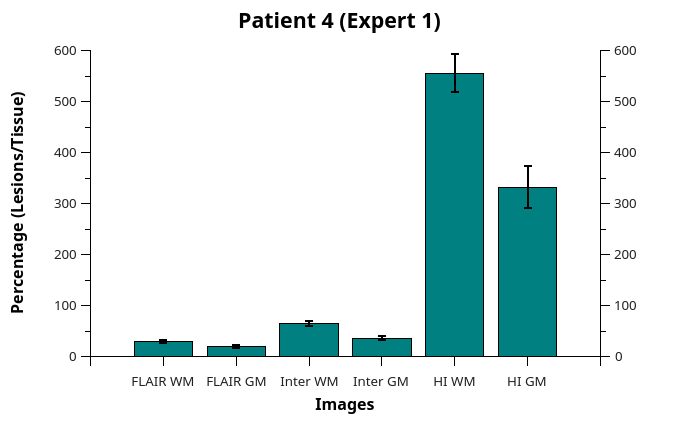}}
\par\end{centering}
\begin{centering}
\subfloat[]{\centering{}\includegraphics[scale=0.46]{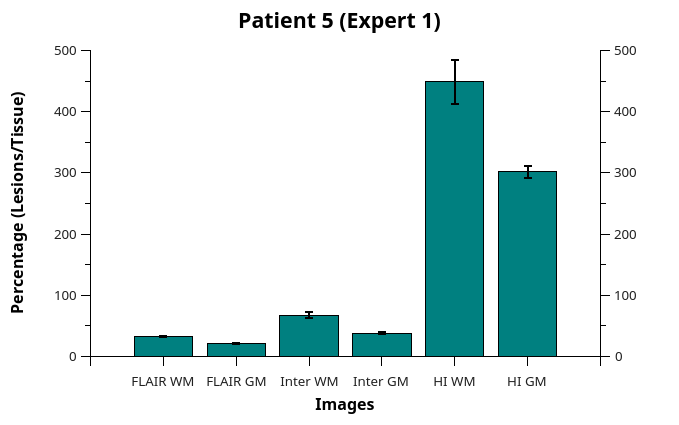}}
\par\end{centering}
\caption{\label{fig:Brightness-expert-1}Lesion intensity profile compared
to white matter and gray matter tissues using ground truths from expert
1. Here, ``Inter'' is the intermediate image and ``HI'' is the
hyperintensity map. Each bar represents the mean lesion intensity
over the mean intensity of a given tissue (white matter or gray matter)
in a particular image type.}

\end{figure}

\begin{figure}[!tbph]
\subfloat[]{\centering{}\includegraphics[scale=0.46]{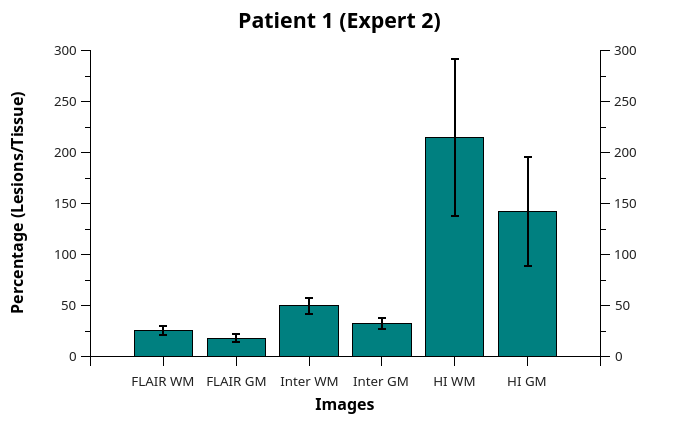}}\subfloat[]{\centering{}\includegraphics[scale=0.46]{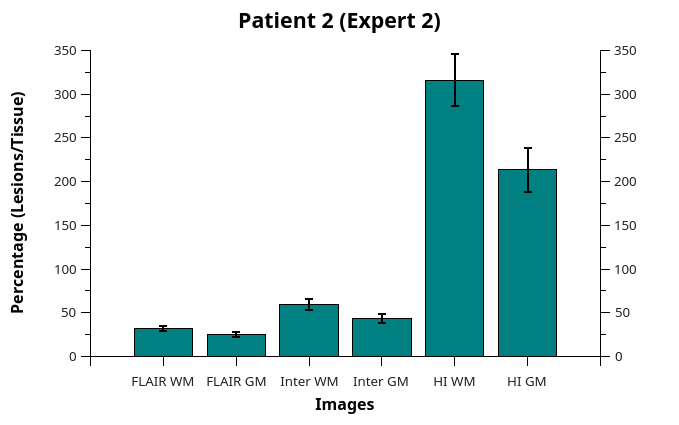}}

\subfloat[]{\centering{}\includegraphics[scale=0.46]{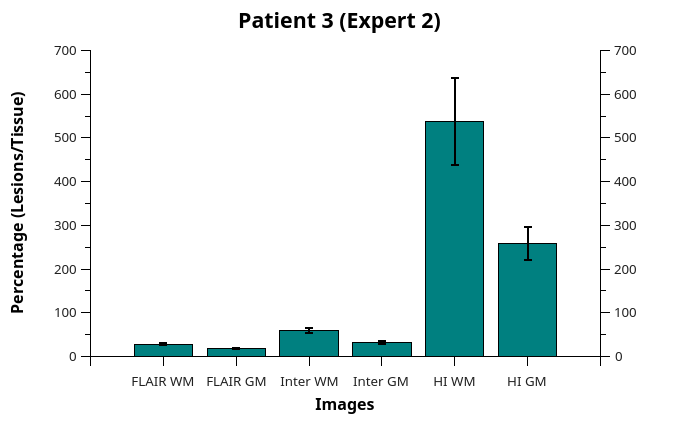}}\subfloat[]{\centering{}\includegraphics[scale=0.46]{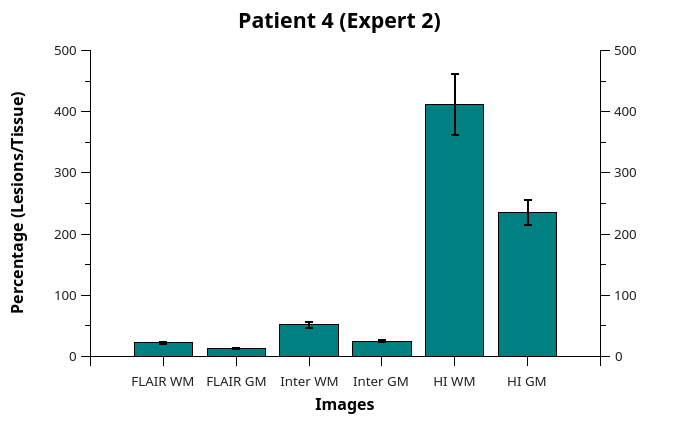}}
\begin{centering}
\subfloat[]{\centering{}\includegraphics[scale=0.46]{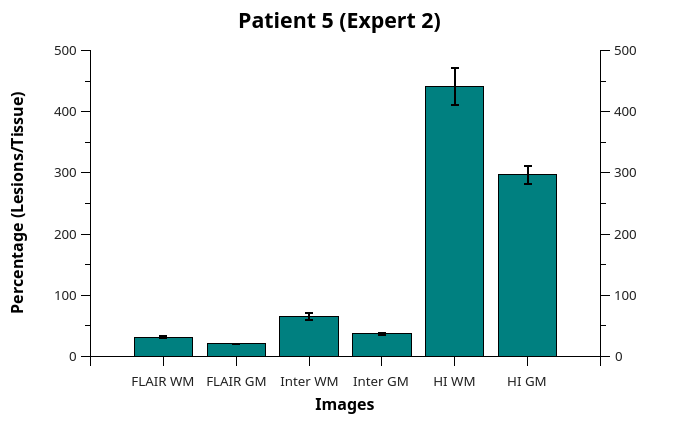}}
\par\end{centering}
\caption{\label{fig:Brightness-expert-2}Lesion intensity profile compared
to white matter and gray matter tissues using ground truths from expert
2. Here, ``Inter'' is the intermediate image and ``HI'' is the
hyperintensity map. Each bar represents the mean lesion intensity
over the mean intensity of a given tissue (white matter or gray matter)
in a particular image type.}
\end{figure}

Since the 2015 Longitudinal MS Lesion Segmentation Challenge provided
two ground truths for each time-point, we extracted the intensity
profiles for both annotations. We rescaled all images to the $[0,1]$
interval, averaged the white matter, gray matter and lesion profiles
and also calculated the standard deviation for each patient. The results
are shown in Figures \ref{fig:Brightness-expert-1} and \ref{fig:Brightness-expert-2}
for experts 1 and 2, respectively.

Each bar in Figures \ref{fig:Brightness-expert-1} and \ref{fig:Brightness-expert-2}
represents the mean lesion intensity over the mean intensity of a
given tissue (gray matter or white matter) in a particular image type
(FLAIR, intermediate and hyperintensity map). For instance, the FLAIR
WM bar in Figure \ref{fig:Brightness-expert-1}(a) must be interpreted
as ``the average MS lesion profile in FLAIR images from Patient 1
is approximately 25\% brighter than the average WM tissue intensity
for the same image type and patient''. This result allows a direct
comparison between tissues and images.

\begin{figure}[h]
\begin{centering}
\subfloat[]{\begin{centering}
\includegraphics[scale=0.6]{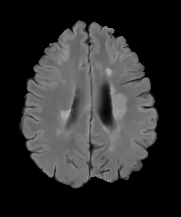}
\par\end{centering}

}\subfloat[]{\begin{centering}
\includegraphics[scale=0.42]{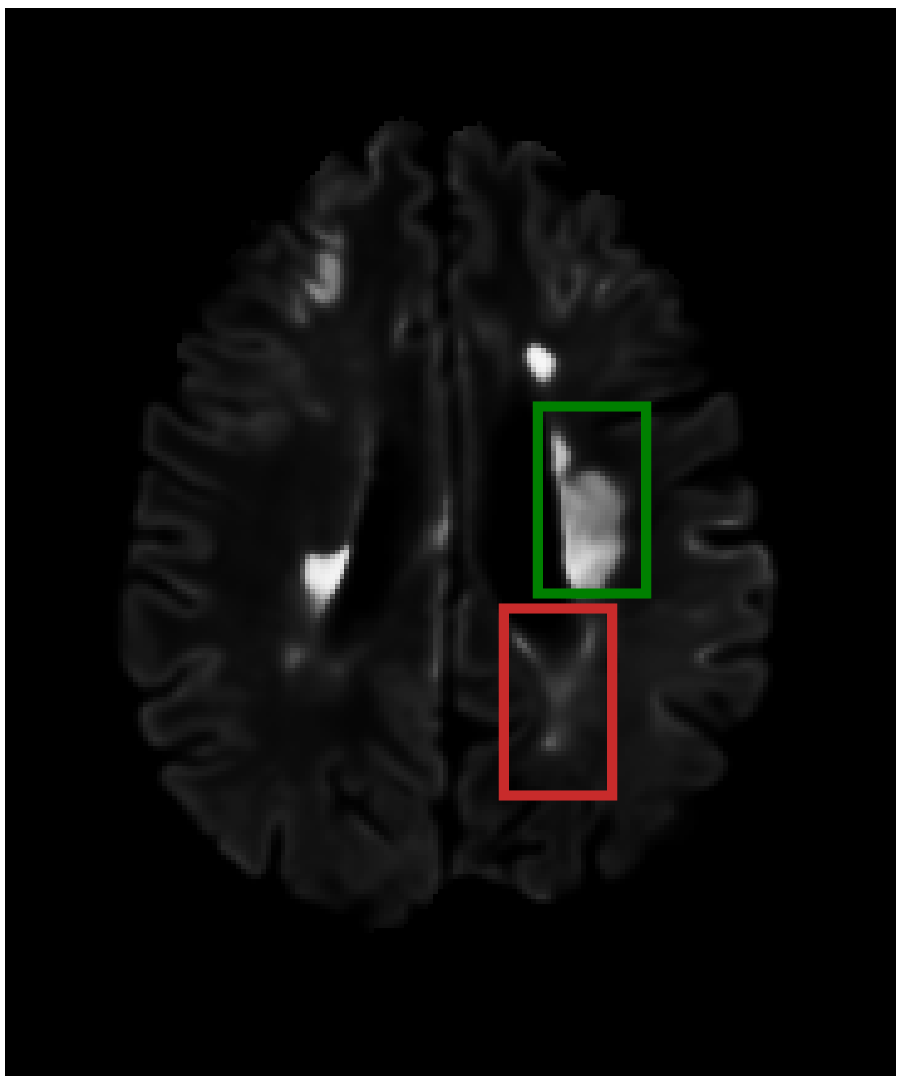}
\par\end{centering}
}\subfloat[]{\begin{centering}
\includegraphics[scale=0.6]{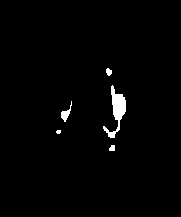}
\par\end{centering}
}
\par\end{centering}
\begin{centering}
\subfloat[]{\begin{centering}
\includegraphics[scale=0.6]{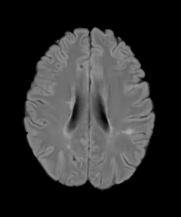}
\par\end{centering}
}\subfloat[]{\begin{centering}
\includegraphics[scale=0.42]{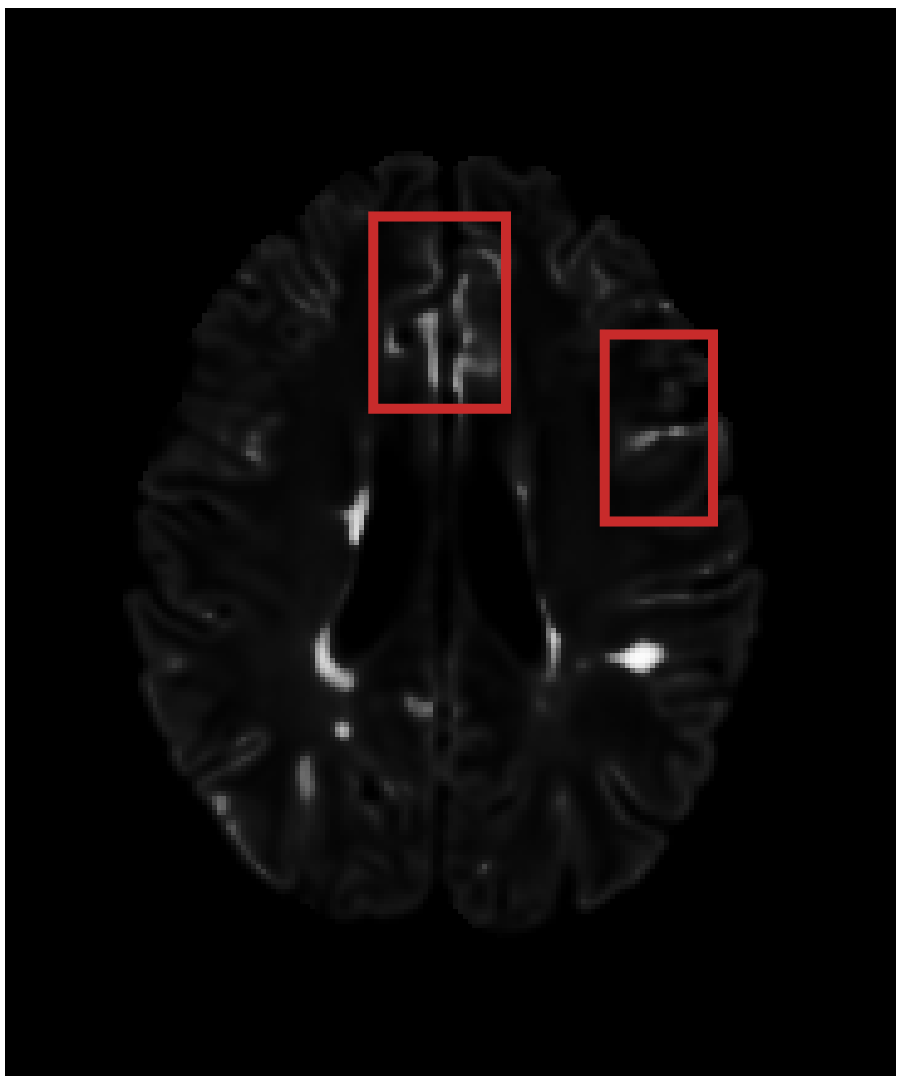}
\par\end{centering}
}\subfloat[]{\begin{centering}
\includegraphics[scale=0.6]{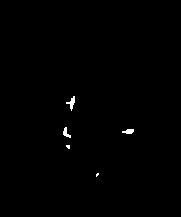}
\par\end{centering}
}
\par\end{centering}
\caption{\label{fig:Intensity-problem}Intensity problems caused by wide lesion
intensity range (first row) and hyperintensities in regions other
than lesions (second row). }

\end{figure}

\subsection{\label{subsec:White-matter-comparison}White matter mask comparison}

\begin{figure}[!tbph]
\begin{centering}
\includegraphics[scale=0.6]{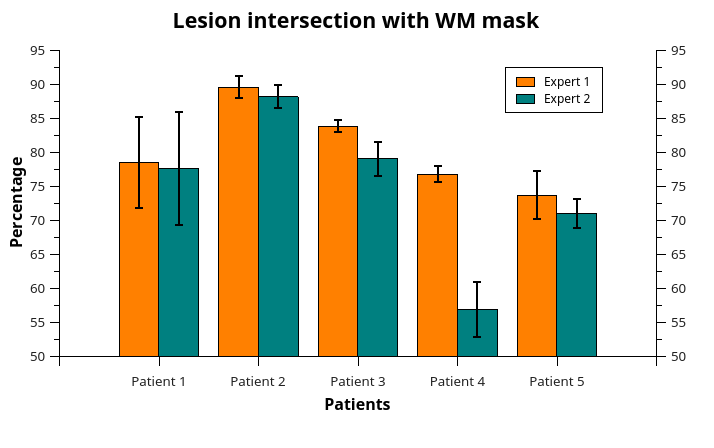}
\par\end{centering}
\caption{\label{fig:Intersection-graph}Lesion intersection with the estimated
white matter mask using ground truths from both experts.}
\end{figure}

\begin{figure}[!tbph]
\begin{centering}
\includegraphics[scale=0.6]{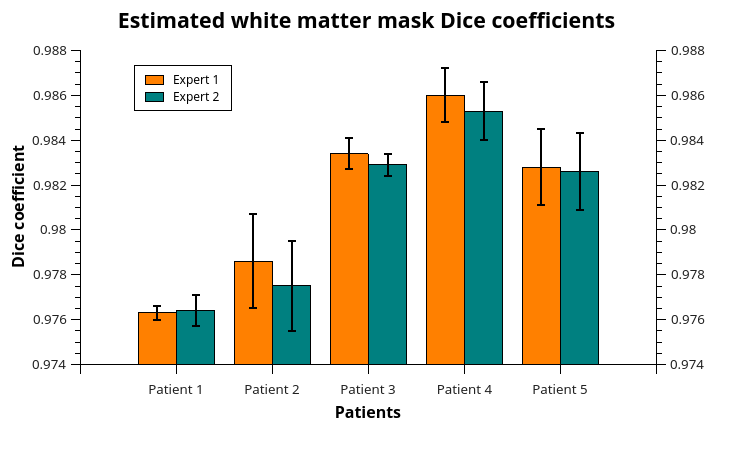}
\par\end{centering}
\caption{\label{fig:Dice-graph}Dice coefficients of the estimated white matter
mask compared to ground truths created using experts lesion annotations
and automatic brain segmentation.}
\end{figure}

We compared the white matter mask estimated in Section \ref{subsec:White-matter-mask}
with our WM ground truths using the Dice coefficient. We also extracted
the percentage of lesions (intersection) present in each estimated
mask to analyze the lesion load that was kept during the estimation.
Again, since there were two lesion ground truths for each time-point,
we extracted metrics for both experts. The results are presented in
Table \ref{tab:Intersection-Dice-Experts} and shown in Figures \ref{fig:Intersection-graph}
and \ref{fig:Dice-graph}. 

\begin{table}[h]
\begin{centering}
\begin{tabular}{c>{\centering}p{2cm}>{\centering}p{2cm}>{\centering}p{2cm}>{\centering}p{2cm}}
\toprule 
 & LI (\%) 

Expert 1

($\mu\pm\sigma$) & LI (\%)

Expert 2 ($\mu\pm\sigma$) & Dice

Expert 1

($\mu\pm\sigma$) & Dice 

Expert 2

($\mu\pm\sigma$)\tabularnewline
\midrule
\midrule 
Patient 1 & 78.56 $\pm$ 6.7  & 77.65 $\pm$ 8.30 & \begin{centering}
0.9763 $\pm$
\par\end{centering}
\centering{}0.0003  & \centering{}0.9764 $\pm$ 0.0007 \tabularnewline
\midrule
\midrule 
Patient 2 & 89.60 $\pm$ 1.59 & 88.20 $\pm$ 1.71  & \centering{}0.9786 $\pm$ 0.0021 & \centering{}0.9775 $\pm$ 0.0020\tabularnewline
\midrule
\midrule 
Patient 3 & 83.88 $\pm$ 0.94 & 79.08 $\pm$ 2.51  & \centering{}0.9834 $\pm$ 0.0007  & \centering{}0.9829 $\pm$ 0.0005 \tabularnewline
\midrule
\midrule 
Patient 4 & 76.83 $\pm$ 1.20  & 56.95 $\pm$ 4.03  & \centering{}0.9860 $\pm$ 0.0012 & \centering{}0.9853 $\pm$ 0.0013 \tabularnewline
\midrule
\midrule 
Patient 5 & 73.73 $\pm$ 3.55 & 71.00 $\pm$ 2.16  & \centering{}0.9828 $\pm$ 0.0017 & \centering{}0.9826 $\pm$ 0.0017 \tabularnewline
\bottomrule
\end{tabular}
\par\end{centering}
\caption{\label{tab:Intersection-Dice-Experts}Lesion intersection (LI) and
Dice coefficients for the white matter mask estimation for both expert
ground truths.}
\end{table}

\section{\label{sec:Discussion}Discussion}

The results in Figures \ref{fig:Brightness-expert-1} and \ref{fig:Brightness-expert-2}
indicate a significant difference in the MS lesion intensity profile
in the hyperintensity map compared to FLAIR and the intermediate image
described in Section \ref{subsec:Pre-processing}. This result is
a significant outcome, because it provides quantitative background
to show the discriminative features of the HI map. 

It is possible to note that the lesion intensity profile was more
similar to gray matter than to white matter. The difference in intensity
between MS lesions and these two tissues in FLAIR and intermediate
images was minimal compared to the HI map, which showed, in the worst
case (Patient 1, Expert 2, GM), a 141\% brightness gap. In contrast,
this same case presented an 18\% and 32\% IPD for FLAIR and intermediate
images. 

At the same time, the standard deviation in the HI map was far bigger
than in FLAIR and intermediate images, which is an indicator that
the map has a rather spread out MS lesion intensity profile. While
this is a concern that must be addressed when using the map to segment
lesions, whether manually or automatically, the overall difference
in intensity between lesions and other brain tissues is still significant
and may provide enough leverage to overcome, at least partially, the
wide standard deviation variation.

Moreover, a distinct drawback of the HI map is that it is highly dependent
on the gray level intensity in FLAIR. This fact poses two problems
that can be seen in Figure \ref{fig:Intensity-problem}. The first
one is presented in Figure \ref{fig:Intensity-problem}(a)-(c) and
concerns the natural intensity variation within the lesion profile.
In this case, the lesion enclosed by the red rectangle was not as
enhanced as the one enclosed by the green rectangle. As seen in Figure
\ref{fig:Intensity-problem}(a), the red rectangle lesion did not
present a profile as hyperintense as the green one in FLAIR, so this
difference was propagated to the HI map. The other problem is shown
in Figure \ref{fig:Intensity-problem}(d)-(f). Areas enclosed by red
rectangles indicate regions that do not have MS lesions and yet are
enhanced in the map. Again, this happens because these regions presented
rather high-intensity profiles in FLAIR and thus were enhanced in
the map. Both these problems may interfere with lesion segmentation
accuracy and indicate that the HI map should not be used as a stand-alone
feature in manual and automatic segmentation techniques.

While there are some works in the literature \cite{Shah-2011,Tomas-Fernandez-2015,Battaglini-2012}
that mention MS lesion intensity profiles and how they relate to other
tissues, none of them provide a quantitative analysis regarding percentage,
making it difficult to to compare our results with theirs objectively.
The databases are also different. However, analyzing scatter plots
in \cite{Shah-2011,Tomas-Fernandez-2015}, it is possible to see that
lesion intensity profiles present a significant overlap with other
brain tissues. Hence, the HI map can undoubtedly help distinguish
lesions more easily.

On the white matter mask estimation, the results are shown in Table
\ref{tab:Intersection-Dice-Experts} and in Figures \ref{fig:Intersection-graph}
and \ref{fig:Dice-graph} indicate high DSCs and a significant intersection
with lesions. A relevant observation to be made is that the LI metric
presented a consistent level of intersection regardless of lesion
volumes, which is an indication of robustness. 

In Figure \ref{fig:Intersection-graph}, it is possible to note that
patient 4 presented very different results on lesion intersection.
The reason for this is that the expert annotations for this patient
had the smallest DSC ($0.612\pm0.0019$) among all patients, as mentioned
in \cite{Freire-2016}. In other words, experts did not have a high
agreement coefficient on lesion segmentation for this particular case,
which consequently made our technique present very different intersection
values for each annotation. 

Another point to be made about Figure \ref{fig:Intersection-graph}
is that patient 1 presented the biggest standard deviation of all.
This result came from the fact that this patient's lesion intensities
faded across time-points, making the enhancement less effective. This
fading phenomenon can also be seen in Figures \ref{fig:Brightness-expert-1}
and \ref{fig:Brightness-expert-2}, since patient 1 had the biggest
standard deviation on the lesion intensity profile in the HI map compared
to other patients.

As mentioned in Section \ref{sec:Introduction}, there are various
works in the literature focused on automatic brain tissue segmentation
\cite{Battaglini-2012,Valverde-2015,Valverde-2015b,Valverde-2016}.
Though a direct comparison is not possible due to different metrics
being used and database access restrictions, the closest work to ours
regarding white matter estimation was \cite{Valverde-2016}. Contrary
to the authors approach, our proposal requires only FLAIR images (from
which the HI maps are created) and WM probability atlases to fill
lesion holes left out during an automatic segmentation process. It
is important to note that our technique focuses only on the white
matter region at this point, while all three major brain tissues are
segmented in \cite{Valverde-2016}. However, we believe that it is
possible to extend our work to also handle gray matter and cerebrospinal
fluid tissues, thus allowing a thorough comparison between techniques
in the future.

\section{\label{sec:Conclusions}Conclusions}

This work presented an automatic technique based on the works of \cite{Roy-2013,Roy-2014}
to enhance hyperintensities in FLAIR images, making it easier to distinguish
multiple sclerosis lesions from other brain tissues, namely gray matter
and white matter. By defining a metric called Intensity Profile Difference
(IPD), we were able to analyze, percent-wise, how much brighter the
lesion profile was compared to other tissues and image types on five
patients from the 2015 Longitudinal Multiple Sclerosis Lesion Segmentation
Challenge. 

The hyperintensity map, created by the enhancement process, offered
a much more distinct lesion profile compared to FLAIR. On average,
lesions presented a mean intensity profile 444.57\% and 264.88\% brighter
than white matter and gray matter in the HI map, respectively. In
FLAIR, the same profile was only 25\% and 19\% brighter considering
the same tissues. This result may serve as an essential aid for both
manual and automatic segmentation techniques.

The HI map is heavily based on gray level intensities and has two
drawbacks worth of notice. The first one regards the intensity variation
within lesions since the difference in intensity from one lesion to
another in FLAIR is propagated to the map. The second drawback regards
regions that do not have lesions but also appear hyperintense in FLAIR.
These regions are also enhanced in the map, which may lead to false
positives. Due to this reason, the HI map should not be used as a
stand-alone feature in applications such as tissue or lesion segmentation.

A byproduct of the HI map is an initial estimate of the white matter
mask for a given time-point. Automatically segmenting a brain image
with multiple sclerosis into three clusters (white matter, gray matter,
and cerebral spinal fluid) will undoubtedly mix lesions with other
tissues. In this sense, we can simply get the white matter cluster
mask and fill regions that are not yet in the mask but are above a
certain threshold in the map to get a ``full'' WM mask. The estimation
of the white matter area may be relevant to narrow down the ROI when
segmenting lesions and also help with brain tissue segmentation and
volume assessment.

We showed that lesions have an intensity profile that is brighter
than white matter than it is to gray matter, so we believe that restricting
the segmentation area to a mask that excludes most gray matter region
might increase lesion segmentation accuracy. However, the two problems
mentioned before about the HI map also affected the white matter mask
estimation. Part of the lesions was left out, as evidenced by the
LI metric, and some regions not related to white matter were included
in the estimation. 

In conclusion, the results of this study showed that the hyperintensity
map offers a much more distinct profile for multiple sclerosis lesions
compared to white matter and gray matter tissues in FLAIR and such
map can also be used to estimate an initial white matter mask. In
future works, we aim to address the problems with the enhancement
algorithm mentioned earlier and efficiently use it to increase both
tissue and lesion segmentation accuracies in automatic techniques.
We also plan on extending the white matter estimation technique to
the other two primary brain tissues (white matter and cerebrospinal
fluid).

\section*{Acknowledgments}

The authors would like to thank the São Paulo Research Foundation
(FAPESP) for the financial support given to this research (grant 2016/15661-0).

\bibliographystyle{plain}
\bibliography{MSBibliography}

\end{document}